\documentclass[conference]{IEEEtran}
\IEEEoverridecommandlockouts
\usepackage{cite}
\usepackage{amsmath,amssymb,amsfonts}
\usepackage{algorithmic}
\usepackage{graphicx}
\usepackage{textcomp}
\usepackage{xcolor}

\def\BibTeX{{\rm B\kern-.05em{\sc i\kern-.025em b}\kern-.08em
    T\kern-.1667em\lower.7ex\hbox{E}\kern-.125emX}}
\begin{document}

\title{Invariant Filtering for Bipedal Walking\\
on Dynamic Rigid Surfaces with Orientation-based Measurement Model

%
}

\author{\IEEEauthorblockN{Yuan Gao}
\IEEEauthorblockA{\textit{Department of Mechanical Engineering} \\
\textit{University
of Massachusetts Lowell}\\
Lowell, USA \\
{yuan\_gao}@student.uml.edu}
\and
\IEEEauthorblockN{Yan Gu}
\IEEEauthorblockA{\textit{Department of Mechanical Engineering} \\
\textit{University
of Massachusetts Lowell}\\
Lowell, USA  \\
{yan\_gu}@uml.edu}

}

\maketitle

\section{Introduction}
\vspace{-0.05in}
Real-world applications of bipedal robot walking require accurate, real-time state estimation. 
State estimation for locomotion over dynamic rigid surfaces (DRS)~\cite{iqbal2020provably}, such as elevators, ships, public transport vehicles, and aircraft, remains under-explored, although state estimator designs for stationary rigid surfaces have been extensively studied~\cite{bloesch2013state,bledt2018cheetah,hartley2020contact}.
Addressing DRS locomotion in state estimation is a challenging problem mainly due to the nonlinear, hybrid nature of walking dynamics~\cite{gao2019dscc,KONG2021109752}, the nonstationary surface-foot contact points~\cite{iqbal2021modeling}, and hardware imperfections (e.g., limited availability, noise, and drift of onboard sensors).

Towards solving this problem, we introduce an Invariant Extended Kalman Filter (InEKF) whose process and measurement models explicitly consider the DRS movement and hybrid walking behaviors while respectively satisfying the group-affine condition and invariant form.
Due to these attractive properties, the estimation error convergence of the filter is provably guaranteed for hybrid DRS locomotion.
The measurement model of the filter also exploits the holonomic constraint associated with the support-foot and surface orientations, under which the robot's yaw angle in the world becomes observable in the presence of general DRS movement.
Experimental results of bipedal walking on a rocking treadmill demonstrate the proposed filter ensures the rapid error convergence and observable base yaw angle.


\vspace{-0.05in}
\section{Method}
\vspace{-0.05in}

In this study, we provably expand the existing invariant extended Kalman filter (InEKF) for legged locomotion on stationary surfaces~\cite{hartley2020contact}
to explicitly addressing the discrete walking behaviors (e.g., support-foot switching) and the time-varying DRS movement.
We choose to build the proposed filter design upon the InEKF framework~\cite{7523335} because of its attractive properties in guaranteeing error convergence.
Specifically, the InEKF ensures that the error dynamics are independent from the estimated state when the process and measurement models are respectively group-affine and invariant, thus leading to the provable error convergence even under large initial errors.
In addition, we form a new, invariant measurement model based on the alignment of the orientations of the support foot and flat DRS.
Combined with the position-based invariant measurement model in~\cite{hartley2020contact}, which is omitted for brevity, the proposed orientation-based measurement model can render the base yaw angle observable as long as the DRS is flat and does not remain perpendicular to the gravity. 
The proposed filter design is briefly explained next.

\begin{figure}[t]
    \centering
    \includegraphics[width=0.95\linewidth]{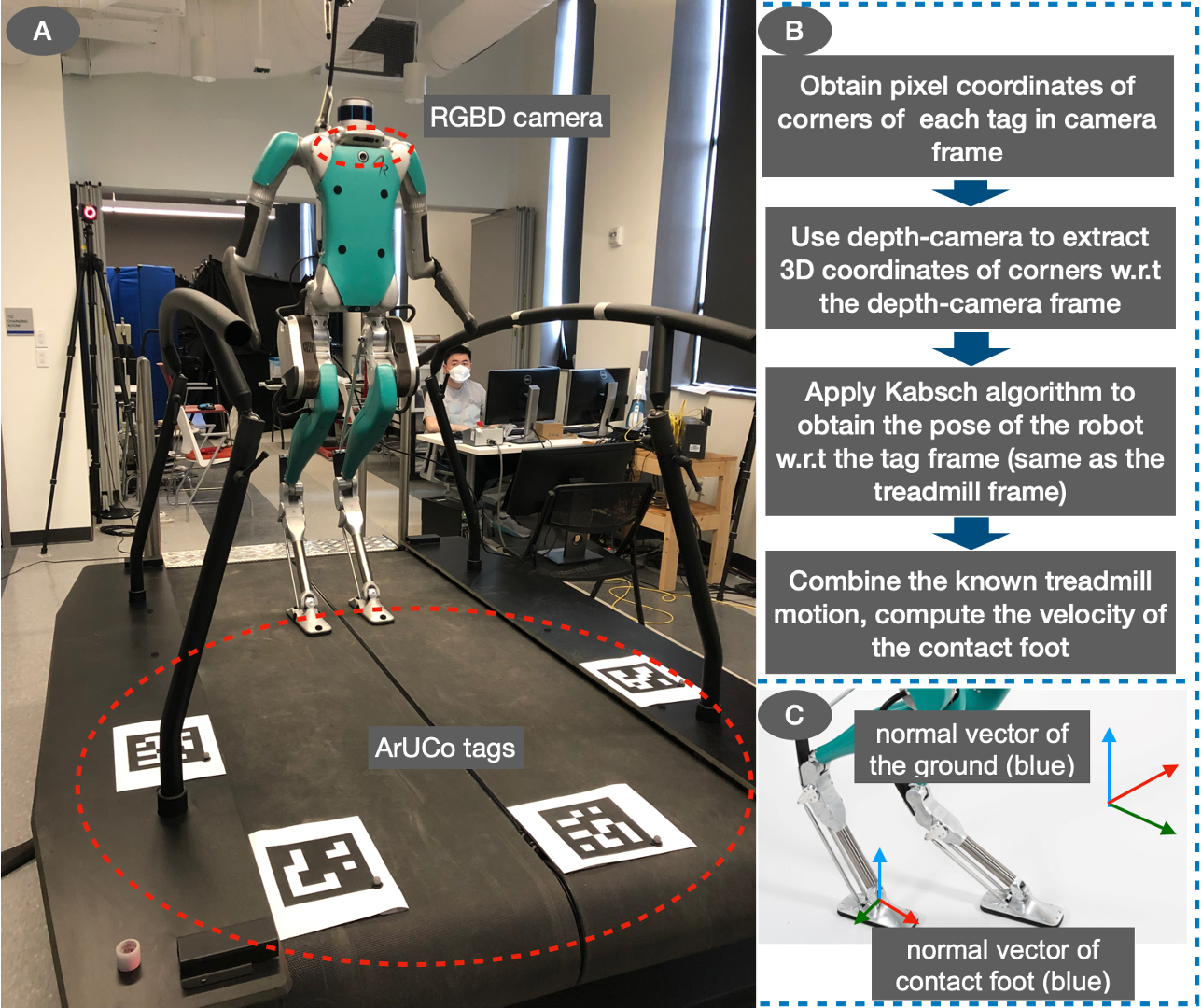}
    \caption{A: Experimental setup.
    B: Procedure of obtaining the support-foot velocity with Digit's RGBD camera, ArUCo tags, and known treadmill motion.
    C: Aligned normal vectors to the surface and support foot.}
    \label{Fig:experiment_setup}
    \vspace{-0.2in}
\end{figure}

\vspace{-0.05in}
\subsection{Propagation for the Continuous-Time Subsystem}

We choose to estimate the following variables:
base position $\mathbf{p}_t=[p_x,p_y, p_z]^T$,
base velocity $\mathbf{v}_t = [v_x,v_y,v_z]^T$,
base orientation $\mathbf{R}_t\in SO(3)$,
and support-foot position $\mathbf{d}_t=[d_x,d_y,d_z]^T$.
These variables are commonly used in motion planning and controller design for legged locomotion.

The state is expressed on the matrix Lie group $SE_3(3)$ as:
\vspace{-0.05in}
\begin{equation}
\small
\label{equ: sim state}
\begin{aligned}
    \mathbf{X}_t &:=
    \begin{bmatrix}
    \mathbf{R}_t& \mathbf{v}_t & \mathbf{p}_t & \mathbf{d}_{t} 
    \\
    \mathbf{0}_{1 \times 3} & 1 &0&0
    \\
    \mathbf{0}_{1 \times 3} & 0 &1&0
    \\
    \mathbf{0}_{1 \times 3} & 0 &0&1
    \end{bmatrix}
    \in SE_3(3)
\end{aligned}
\vspace{-0.05in}
\end{equation}
\vspace{-0.05in}

The input $\mathbf{u}_t$ is defined as
$
    \mathbf{u}_t:=
    [
    \Tilde{\boldsymbol{\omega}}_t
    ,
    \Tilde{\mathbf{a}}_t^T
    ,
    (\tilde{\mathbf{v}}^{d}_t)^T]^T.
$
The vectors $\Tilde{\boldsymbol{\omega}}_t\in\mathbb{R}^3$ and
$\Tilde{\mathbf{a}}_t\in \mathbb{R}^{3}$ are the raw data returned by the gyroscope and accelerometer of the IMU attached to the robot's base (i.e., trunk), respectively.
The vector
$\tilde{\mathbf{v}}^{d}_t \in\mathbb{R}^{3}$ is the linear velocity of the foot-surface contact area obtained using the robot's RGBD camera, the ArUCo tags on the DRS, and the known surface pose in the world.
Here we assume that the surface pose is relatively accurately known since real-world dynamic rigid platforms (e.g., ships) are commonly equipped with motion monitoring systems that return such data. 

The noisy continuous-time process model is given by:
\begin{equation}
\small
\label{equ: sim propagation}
    \begin{aligned}
    \frac{d}{dt}\mathbf{X}_t &= 
    \begin{bmatrix}
    \mathbf{R}_t(\Tilde{{\omega}}_t)_\times & \mathbf{R}_t\Tilde{\mathbf{a}}_t + \mathbf{g} & \mathbf{v}_t & \tilde{\mathbf{v}}^{d}_t
    \\
    \mathbf{0}_{3 \times 3} & \mathbf{0}_{3 \times 1} & \mathbf{0}_{3 \times 1} & \mathbf{0}_{3 \times 1}
    \end{bmatrix}
    +\mathbf{X}_t\boldsymbol{w}_t,
    \\
    &=\mathbf{f}_{u_t}(\mathbf{X}_t)+\mathbf{X}_t \boldsymbol{w}_t
    \end{aligned}
\end{equation}
with $\boldsymbol{w}_t:=([\mathbf{w}^g_t, (\boldsymbol{w}^a_t)^T, \mathbf{0}_{1 \times 3}, (\boldsymbol{w}^d_t)^T]^T)^\wedge$. 
The vectors $\mathbf{w}^g_t$, $\boldsymbol{w}^a_t$, and $\boldsymbol{w}^d_t$ are continuous Gaussian white noises associated with the measured angular velocity and linear acceleration of the base as well as the contact-velocity measurement, respectively.
The vector $\mathbf{g}$ is the gravitational acceleration.
It can be proved that $\mathbf{f}_{u_t}$ meets the group-affine property~\cite{7523335}.

\vspace{-0.05in}
\subsection{Measurement Update for the Continuous-Time Subsystem}

When the support foot remains a secured contact with the DRS, the normal vector to the surface and that to the support foot are parallel (see Fig.~\ref{Fig:experiment_setup}C).
Based on this kinematic relationship, we introduce a new measurement model in the right-invariant form as follows:
\vspace{-0.05in}
\begin{equation}
\label{equ: right inv foot orientation}
\small
\begin{aligned}
    \underbrace{\begin{bmatrix}
    ^b\mathbf{R}_f
    \begin{bmatrix}
        \mathbf{0}_{2 \times 1}
        \\
        1
    \end{bmatrix}
    \\
    \mathbf{0}_{3 \times 1}
    \end{bmatrix}}_{\mathbf{Y}_{t}} =
    \mathbf{X}_t^{-1}
    \underbrace{\begin{bmatrix}
    \mathbf{R}_s
    \begin{bmatrix}
    \mathbf{0}_{2 \times 1}
        \\
        1
    \end{bmatrix}
    \\
    \mathbf{0}_{3 \times 1}
    \end{bmatrix}}_{\mathbf{d}_{t}}
    +
    \mathbf{V}_{t},
\end{aligned}
\vspace{-0.05in}
\end{equation}
where $\mathbf{R}_s\in SO(3)$ is the orientation of DRS in the world frame,
$^b\mathbf{R}_f\in SO(3)$ is the orientation of the support foot in the base frame,
and
$\mathbf{V}_{t}$ is the noise term accounting for the uncertainty in the knowledge of the surface orientation.

Besides the orientation-based measurement model, the proposed filter also includes the right-invariant, position-based measurment model as in~\cite{hartley2020contact}.
Following the standard steps of the InEKF methodology~\cite{7523335}, we obtain the equations for the update step of the filter, which are omitted for brevity. 

\begin{figure}[t]
    \centering
    \includegraphics[width=0.9\linewidth]{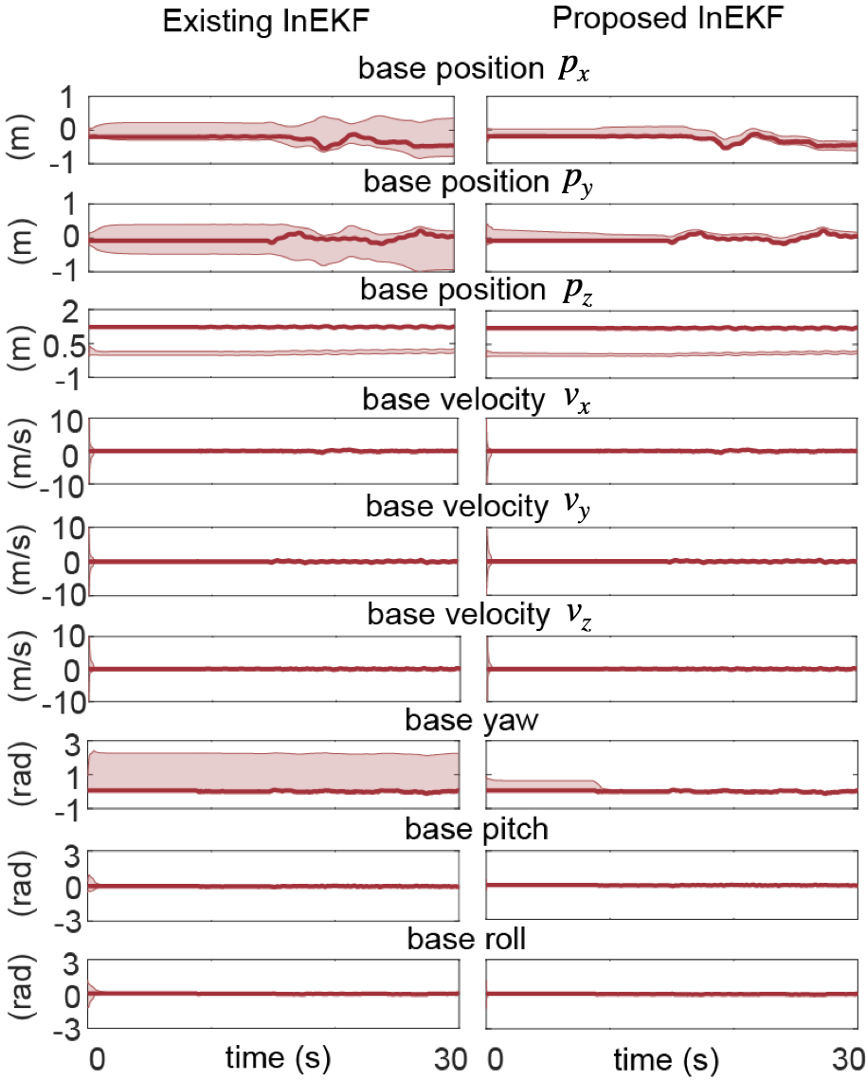}
    \caption{Estimation results of the existing~\cite{hartley2020contact} and proposed InEKF designs.}
    \label{Fig:100_comparison}
    \vspace{-0.2in}
\end{figure}

\vspace{-0.05in}
\subsection{Propagation at Jumps}

When the swing foot strikes the DRS, the support and swing legs switch roles, causing a sudden jump in the support-foot position $\mathbf{d}$. 
We express the process model of the jump as:
\vspace{-0.05in}
\begin{equation}
\label{equ: sim jump}
\small
\begin{aligned}
    {\mathbf{X}_{t^+}} &= \begin{bmatrix}
    \mathbf{R}_{t} & \mathbf{v}_{t} & \mathbf{p}_{t} & {\mathbf{d}_{t}
    + \mathbf{R}_{t} \mathbf{h}_d(\mathbf{\Tilde{q}}_t)}
    \\
    \mathbf{0}_{1\times2} & 1 &0&0
    \\
    \mathbf{0}_{1\times2} & 0 &1&0
    \\
    \mathbf{0}_{1\times2} & 0 &0&1
    \end{bmatrix}
    \text{exp}(\boldsymbol{w}^{\Delta}_t)
    ,
\end{aligned}
\end{equation}
\vspace{-0.05in}
where the noise vector $\boldsymbol{w}^{\Delta}_t \in \mathbb{R}^{\text{dim} \boldsymbol{\mathfrak{g}}}$ is induced by the robot's encoder noise.
The variable $\mathbf{h}_d$ is the
new support-foot position just after a jump relative to the previous support-foot position with respect to the base frame.
The input to the jump map is the joint angle data $\mathbf{\Tilde{q}}_t$ provided by encoders. 

Since the jump map meets the identity jump map condition for right-invariant errors in~\cite{yuan2021MECC},
the right-invariant error does not jump.
Hence, the Jacobian matrix of the jump map is identity. 
If we use the Jacobian to propagate the estimated covariance, then the estimated covariance does not jump either.

\vspace{-0.05in}
\section{Results}
\vspace{-0.05in}
Figure~\ref{Fig:experiment_setup}A shows the overall experiment setup, which comprises a) a Motek instrumented treadmill with a time-varying pitch angle $\theta_{DRS}= 3^{\circ}sin(1.5\pi t)$, b) the Digit bipedal humanoid robot (developed by Agility Robotics), c) four ArUCo tags attached to the treadmill surface, and d) a motion capture system that records the ground truth.
Digit's on-board sensors used by the proposed filter are: a) IMU at the base, b) joint encoders, and c) RGBD camera at the neck.
The filter also utilizes Digit's default contact estimator to detect foot-landing events.
The procedure to obtain the support-foot velocity $\tilde{\mathbf{v}}^{d}_t$ is shown in Fig.~\ref{Fig:experiment_setup}B.


Figure~\ref{Fig:100_comparison} shows the estimation results from 100 trials of the previous position-based InEKF~\cite{hartley2020contact} and our proposed filter. 
The results show that both filter achieve rapid error convergence for the linear velocity and roll and pitch angles of the base, even in the presence of large initial errors.
Although the base position is unobservable under both filters, the proposed filter seems to maintain a relatively smaller error within the tested walking period (i.e., 30 s).
More importantly, thanks to the orientation-based measurement model, the proposed filter ensures that the base yaw angles is observable under the rocking DRS movement.
In contrast, the yaw angle remains unobservable under the previous InEKF design, as indicated by its large error throughout the entire tested period.


\bibliography{IEEEexample}

\bibliographystyle{ieeetr}

\end{document}